\pdfoutput=1

\documentclass[11pt]{article}

\usepackage[final]{acl}

\usepackage{times}
\usepackage{latexsym}

\usepackage[T1]{fontenc}

\usepackage[utf8]{inputenc}

\usepackage{microtype}

\usepackage{amsmath, bm}
\usepackage{graphicx}
\usepackage{multicol}  
\usepackage{multirow} 

\usepackage{makecell}

\usepackage{xcolor}
\definecolor{c1}{RGB}{220, 28, 19}
\definecolor{c2}{RGB}{234, 76, 70}
\definecolor{c3}{RGB}{240, 116, 112}
\definecolor{c4}{RGB}{241, 149, 155}
\definecolor{c5}{RGB}{246,189,192}

\usepackage{tikz}
\newcommand*\circled[1]{\tikz[baseline=(char.base)]{
            \node[shape=circle,draw,inner sep=0.8pt] (char) {#1};}}
\usepackage{amssymb}
\usepackage{amsfonts}
\usepackage{amsmath}
\usepackage{subfig}
\DeclareMathOperator*{\argmax}{arg\,max}

\title{Parameter-Efficient Tuning by Manipulating Hidden States of Pretrained Language Models For Classification Tasks}

\author{Haoran Yang\textsuperscript{1} , Piji Li\textsuperscript{2},Wai Lam\textsuperscript{1}   \\
  \textsuperscript{1}The Chinese University of Hong Kong \\
  \textsuperscript{2}Tencent AI Lab \\ 
  \texttt{\{hryang,wlam\}@se.cuhk.edu.hk} \\
  \texttt{lipiji.pz@gmail.com} \\
}

\begin{document}
\maketitle
\begin{abstract}
Parameter-efficient tuning aims to distill knowledge for downstream tasks by optimizing a few introduced parameters while freezing the pretrained language models (PLMs). Continuous prompt tuning which prepends a few trainable vectors to the embeddings of input  is one of these methods and  
has drawn much attention due to its effectiveness and efficiency. This family of methods can be illustrated as exerting nonlinear transformations of hidden states inside PLMs.
However, a natural question is ignored: can the hidden states be directly used for classification without changing them? In this paper, we aim to answer this question by proposing a simple tuning method which only introduces three trainable vectors. Firstly, we integrate all layers hidden states using the introduced vectors. And then, we input the integrated hidden state(s) to a task-specific \textit{linear classifier} to predict categories. This scheme is similar to the way ELMo utilises hidden states except that they feed the hidden states to LSTM-based models. Although our proposed tuning scheme is simple, it achieves comparable performance with prompt tuning methods like P-tuning and P-tuning v2, verifying that original hidden states do contain useful information for classification tasks. Moreover, our method has an advantage over prompt tuning in terms of time and the number of parameters.




\end{abstract}

\section{Introduction}
Pretrained language models (PLMs)~\cite{PLM_survey,add1,add2} which are trained on copious amount of corpora have been corroborated valid on an assortment of tasks~\cite{devlin-etal-2019-bert}.
The most commonly adopted method when applying PLMs to downstream tasks is \textbf{fine tuning}, updating all model parameters with labelled data using gradient descent. Although fine tuning is able to obtain good performance, it is unrealistic to store a separate language model for each task during online deployment due to the large size of  current models. For example, RoBERTa-base~\cite{liu2019roberta} and T5-large~\cite{2020t5} have about 125M and 770M parameters respectively. Multi-task learning (MTL)~\cite{liu2019mt-dnn} is a solution to avoid storing multiple copies of LMs. However, in scenarios where the dataset comes in a sequence manner~\footnote{Continual Learning ~\cite{biesialska-etal-2020-continual}  focuses on this scenario.}, MTL is still not an effective strategy.
\begin{figure}[t]
 \centering
  \includegraphics[width=\linewidth]{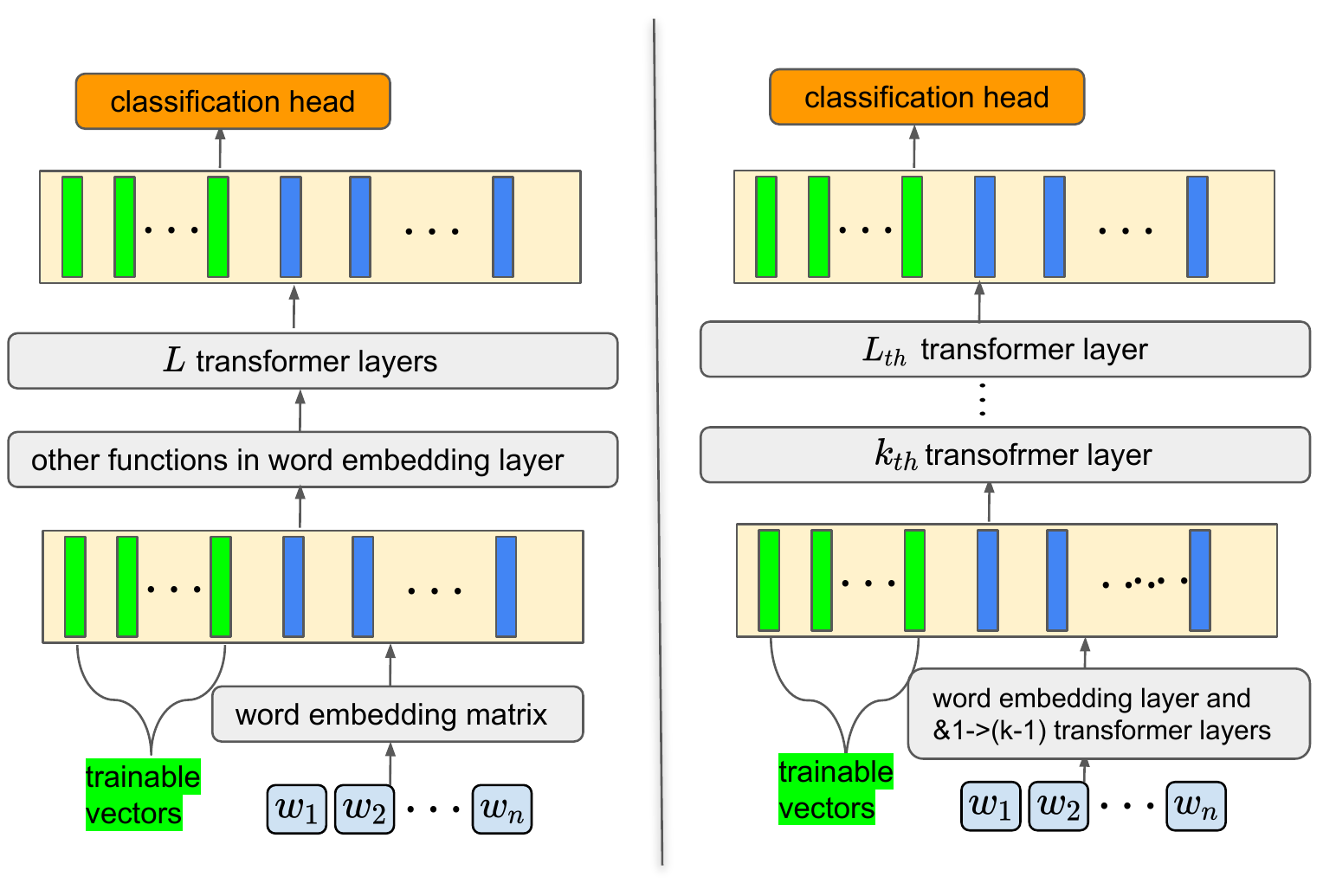}
  \caption{\textbf{Left Figure:} P-tuning~\cite{power_prompt}  prepends some trainable vectors to the word embeddings and forward them to the remaining layers of a PLM. \textbf{Right Figure:} We prepend trainable vectors to the $k_{th}$ layer and find that similar performance can be achieved even when $k=L$ (the final layer).}
  \label{fig:prompt_tuning}
\end{figure}

To reuse PLMs for different tasks, a growing body of \textit{parameter-efficient tuning} methods have been proposed. They 
aim to mine task-specific information from a \textit{frozen} PLM by optimizing only a small amount of extra parameters. Recently the Prompt tuning family methods have received much attention for its superiority in simplicity and parameter efficiency. 
Prompt tuning~\cite{liu_pretrain_prompt} includes discrete prompt tuning and continuous prompt tuning~\footnote{We use continual prompt tuning to represent a family of methods, e.g., P-tuning~\cite{power_prompt}.}.
Discrete prompt tuning ~\cite{NEURIPS2020_1457c0d6,petroni-etal-2019-language,schick-schutze-2021-exploiting} usually inserts some tokens (appearing in the models' vocabulary) to the input sentences to reformulate  tasks as fillin-the-blanks problems. For example, for sentiment classification task where $\bm x$ = ``\textit{This computer is easy to use.}'', the reformulated form of input may be $\bm x'$ =``$\left[ \bm x \right]$, so it is a [MASK] product.''. However, discrete prompt tuning involves huge manual efforts and performs worse on high-resource dataset, limiting its usage. Instead of inserting discrete tokens, continuous prompt tuning~\cite{power_prompt,li-liang-2021-prefix} adds task-specific trainable vectors to the embeddings of input sentences. During training, only these task-specific vectors are optimized. Their experiments show that it is possible to elicit knowledge from frozen PLMs with these extra vectors. 
\begin{figure}[t]
 \centering
  \includegraphics[width=\linewidth]{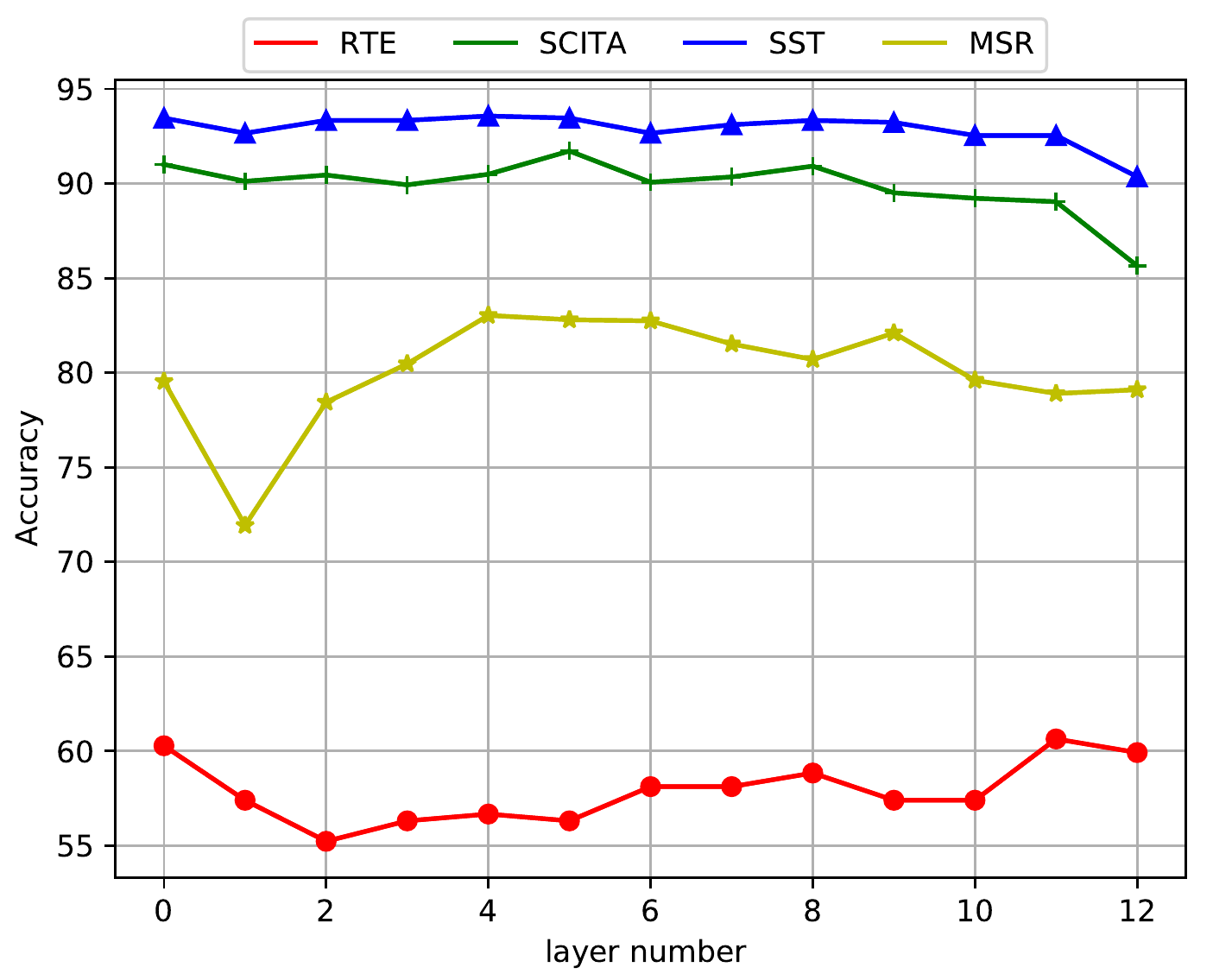}
  \caption{ P-tuning~\cite{power_prompt} prepends trainable vectors to the embedding layer. We prepend trainable vectors to a transformer layer of the RoBERTa-base model. $0$ represents the word embedding layer.}
  \label{fig:layer_acc}
\end{figure}
Different from discrete prompt tuning which aims to take advantage of the masked language model nature by inserting explicit tokens,  
~\citet{power_prompt} find the meaning of inserted vectors is hard to interpret indicating that continuous prompt tuning is completely different from discrete prompt tuning. 
Also,
as shown in Figure~\ref{fig:prompt_tuning}, similar with \textbf{P-tuning} ~\cite{power_prompt} which prepends trainable vectors to the word embeddings,
we attempt to prepend these vectors to a specific transformer layer. Figure~\ref{fig:layer_acc} depicts the accuracy of four tasks. We can see that for each task similar results can be achieved by prepending vectors to any layer, even the final layer ($12_{th}$ layer). Trainable vectors only participate in computations in the final layer if they are prepended to the final layer but thus configuration can still obtain a good performance. 
Does this phenomenon imply that these raw hidden states have already contained most if not all information for classification?


In this paper, we aim to show that \textbf{information for many classification tasks have already been encoded in the raw hidden states of inputs.} To this end,   
we 
propose a simple method which only introduces three trainable vectors to integrate raw hidden states.  
This method firstly obtains all the hidden states of the input and learns a softmax-normalised weight vector to stack hidden states across layers. And secondly, a soft mask vector is used to select a subset of dimensions useful for downstream tasks in the hidden states. Finally, a self-attention operation only composed of one trainable vector is conducted to output the states for classification.
We find that the classification information is on the surface which means that there is no need to use complex classification head like LSTM~\cite{HochSchm97}, CNN. Our classification head only contains a \textit{linear transformation} followed by a \textit{softmax} function.

We conduct experiments on a variety of tasks 
and the results show that our method can attain performance comparable with that achieved by P-tuning~\cite{power_prompt}  and P-tuning v2~\cite{liu2021ptuning}.  
This substantiate that original hidden states do contain necessary information for classification and this kind of information can be extracted with simple linear transformations. \textbf{This finding can also partially explain why p-tuning (v2) works for classification tasks by only introducing a few trainable vectors.}
We hope this finding can help advance the understanding of continuous prompt tuning schemes.

\section{Preliminaries}
We start by introducing some backgrounds of tuning schemes in this section, which describes how ELMo and the transformer family language models are applied to downstream classification tasks.

\subsection{ELMo}
ELMo~\cite{peters-etal-2018-deep}, short for \textbf{E}mbeddings from \textbf{L}anguage
\textbf{Mo}dels, learns contextualized word embedding by pre-training a language model. The backbone of ELMo is $L$ layers of bidirectional LSTMs. After pretraining, when there is a downstream task, ELMo learns a task-specific linear combination of hidden states across layers and inputs these hidden states to LSTM-based models to output a target for an input:
\begin{align}
\label{eq:elmo}
    &\mathbf{e}_k = \gamma \sum_{i=0}^{L}s_j \mathbf{h}_{k,j}\\
    &\hat{y} = Model(\left[\mathbf{e}_1,\mathbf{e}_2,\cdots,\mathbf{e}_n \right])
\end{align}
where $\mathbf{h}_{k,j}$ is the hidden state of the $k_{th}$ token in the $j_{th}$ layer. $\mathbf{h}_{k,0}$ is the representation of the $k_{th}$ token in the word embedding layer.
$(s_0, s_1,\cdots s_L)$ are task-specific softmax-normalized weights and they are optimized when training the downstrem task. These weights allow the model to utilise representations at different levels (e.g., syntax, semantic) based on the properties of the downstream task. $\gamma$ is a trainable scalar which is used to  aid the optimization process. 
$\mathbf{e}_k$ is the final weighted embedding of the $k_{th}$ word. Finally, $\left[\mathbf{e}_1,\mathbf{e}_2,\cdots,\mathbf{e}_n \right]$ are inputted into the tailor-made model (usually a complex network like LSTM) to predict the label or generate the target sentence for a specific task. 
\subsection{Continuous Prompt Tuning}
Large pretrained language models based on transformer~\cite{attnneed} have become the preferred underlying backbone for solving various NLP tasks. When adjusting PLMs to downstream tasks, the most commonly used method is called \textit{fine tuning}, optimizing the PLM and task-specific classifier $C$ together. 
To be more parameter efficient during online deployment, continuous prompt tuning is proposed. There are numerous works towards continuous prompt tuning and we give a detailed introduction to two of them.

\textbf{P-tuning}~\cite{power_prompt} For an input sentence(s) $\bm{x}=(x_1,x_2,\cdots,x_n)$, P-tuning prepends $K$ vectors to the word embedding matrix of $\bm{x}$, denoted by $P(\bm {x}) = \left[\mathbf{w}_0, \mathbf{p}_1,\mathbf{p}_2,\cdots,\mathbf{p}_K,\mathbf{w}_1,\cdots,\mathbf{w}_n 
\right]$. $\mathbf{w}_0$ is the embedding of $\left[CLS\right]$ token. $\mathbf{w}_i$ is the word embedding of the $i_{th}$ token. $\mathbf{p}_i$ is the $i_{th}$ introduced vector. $P(\bm{x})$ is inputted to the remaining layers of a PLM to predict the label of $\bm{x}$. The optimization objective is:
\begin{equation}
\label{pt}
    \min_{\bm{\beta}, \mathbf{p}_{1\cdots K}}{\sum_{\bm{x}\in\bm{D}}{\mathcal{L}(C_{\bm{\beta}}(E(P(\bm{x}))),y)}}
\end{equation}
where $C, E$ represents the task-specific classification head and Encoder (a PLM) respectively. $\bm{\beta}$ represents the parameters of $C$.
We can see that during training, only the prepended vectors and classifiers are optimized. 

\begin{figure*}[t]
 \centering
  \includegraphics[width=0.8\linewidth]{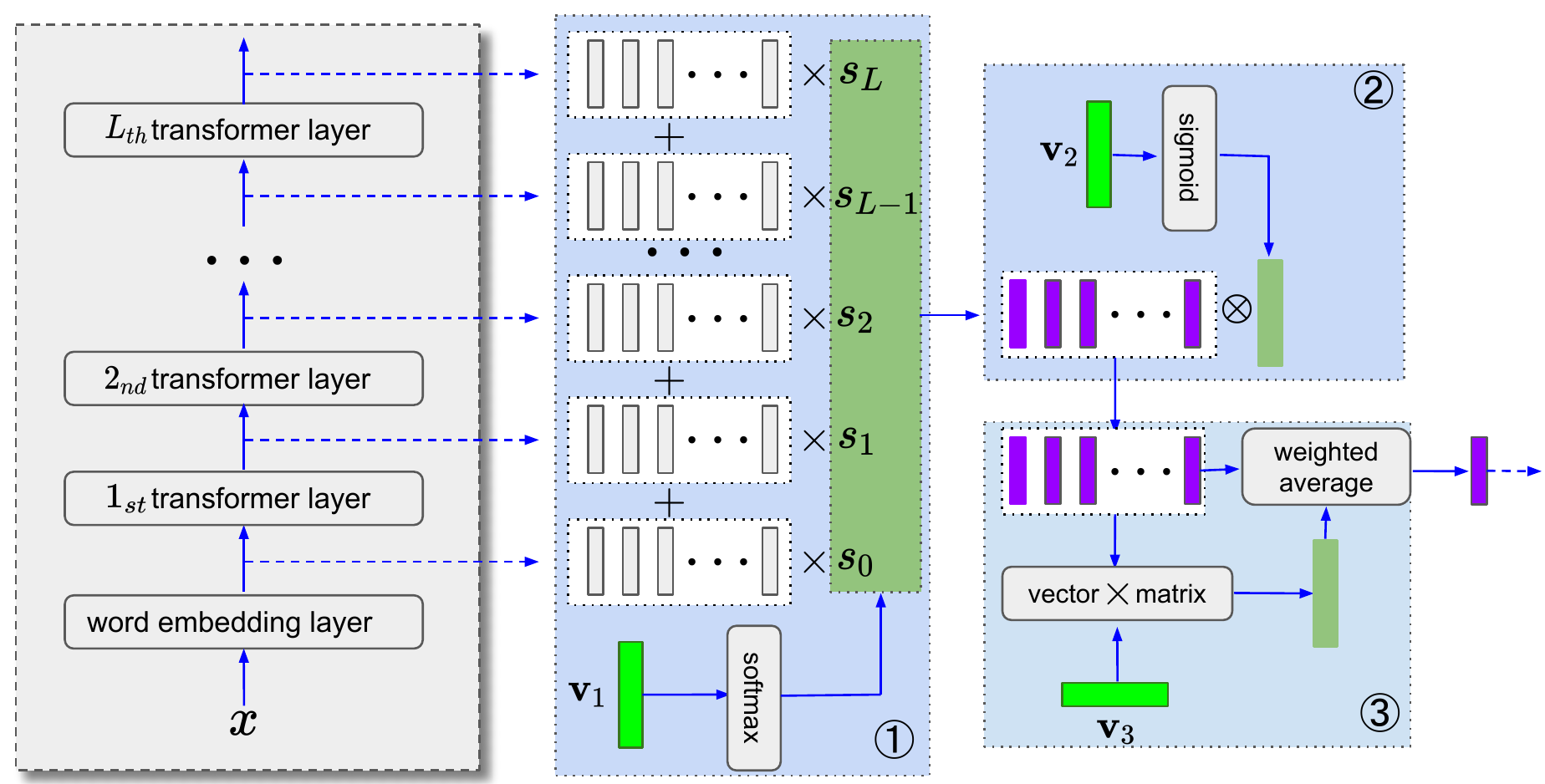}
  \caption{Our proposed framework to distill knowledge from a frozen PLM. \textbf{\circled{\small 1}} Conduct linear combination of hidden states across layers by learning a softmax-normalised weights using $\mathbf{v}_1$. \textbf{\circled{\small 2}} Find the subspace helpful for a specific task in hidden states by learning a soft mask using $\mathbf{v}_2$. \textbf{\circled{\small 3}} Aggregate hidden states by performing self-attention operation using trainable vector $\mathbf{v}_3$. \textbf{\circled{\small 3}} can be omitted if the downstream task is a sequence labeling task.} 
  \label{fig:my_model}
\end{figure*}

\textbf{P-tuning v2 \cite{liu2021ptuning}} is a variant P-tuning which can be scaled to difficult tasks and achieves much better performance compared with P-tuning. Different from P-tuning which inserts vectors to the word embedding matrix and hidden states of these vectors in the following layers are calculated based on the representation in the former layer, P-tuning v2 leverages \textbf{multi-layer prompts} which inserts a set of trainable vectors to each layer of a PLM. In other words, the original hidden states at layer $l$ are affected by prompt vectors in layer $l$ while prompt vectors in layer $l$ is independent with prompt vectors in layer $l-1$. This modification allows for more
per-task capacity since more trainable vectors are introduced.

\section{Proposed Approach}
In this section, we introduce our proposed method which directly manipulates hidden states at different layers. Different from adapters and prompt tuning methods that distill useful information by changing the hidden states, we operate on the raw hidden states. Since there are numerous methods to integrate these hidden states, we only introduce the simplest one that is composed of three parts. 
\paragraph{Layer-Level Weighted Addition} This step is similar to Equation~\ref{eq:elmo} except that we do not use the scalar $\gamma$. Formally, given input  $\bm{x}$, we firstly obtain the hidden states:
\begin{equation}
    \left\{\mathbf{H}_i\right\}_{i=0}^L=  \left\{\mathbf{h}_{i,0}, \cdots,\mathbf{h}_{i,N}\right\}_{i=0}^L = E(\bm{x})
\end{equation}
where $L$ is the number of transformer layers and the $0_{th}$ layer is the word embedding layer, $N$ is the length of $\bm{x}$. Each $\mathbf{h}_{i,j}$ is a $d$-dimensional vector and each $\mathbf{H}_i$ is a $d\times (N+1)$ dimensional matrix. 

Then, we assign each layer a weight since previous works~\cite{jawahar-etal-2019-bert, peters-etal-2018-deep} have shown that low, middle, top layers capture information with different granularity and diverse downstream tasks may pay a different attention to these layers. We achieve this goal by introducing the first trainable vector $\mathbf{v}_1 \in \mathbb{R}^{L+1}$: 
\begin{align}
    & \mathbf{H} = \left\{\mathbf{h}_{0}, \mathbf{h}_{1}, \cdots,\mathbf{h}_{N}\right\} = \sum_{i=0}^L s_i \mathbf{H}_i    \\
    & \mathbf{s} = softmax(\mathbf{v}_1) 
\end{align}
where $s_i$ is the $i_{th}$ item of $\mathbf{s}$.

\paragraph{Subspace Mining}
We argue that for a $d$-dimensional hidden state, there exists a subspace with dimension $d_1$ that contains distinguishing information for a downstream task.  Selecting such subspace with binary mask vector is a choice but it has difficulties about optimization ~\cite{mask}. Here, we adopt the soft mask strategy by introducing the second trainable vector $\mathbf{v}_2 \in \mathbb{R}^d$: 
\begin{align}
    &  \mathbf{H} = \mathbf{m} \otimes \mathbf{H} \\ 
    &\mathbf{m} = sigmoid(\mathbf{v}_2)
\end{align}
where $\otimes$ is the element-wise multiplication. 


\paragraph{Self-Attention}
We use \textit{self-attention} to aggregate the matrix $\mathbf{H}$. There are different forms of self-attention mechanism, for example, the ($\mathbf{Q}$uery, $\mathbf{K}$ey, $\mathbf{V}$alue) self-attention used by transformer ~\cite{attnneed}. But this attention requires a lot of parameters (usually, $\mathbf{Q, K, V}$ are $d\times d$ matrices.). To be more parameter efficient, we use the simplest one which need to introduce only one trainable vector $\mathbf{v}_3 \in \mathbb{R}^d$:
\begin{align}
    & \mathbf{h} = \mathbf{H}\cdot \mathbf{w} \\
    & \mathbf{w} = softmax(\mathbf{v}_3\cdot \mathbf{H}) 
\end{align}

\paragraph{Task-Specific Classification}
With $\mathbf{h}$, we can input them into task-specific classification head. To show that raw hidden states contains most, if not all, information for classification and these information is on the surface,
 we only perform a linear transformation and a softmax operation to output the label: 
\begin{equation}
    y = \argmax (softmax(\mathbf{W}\mathbf{h})) 
\end{equation}
where $\mathbf{W}$ is a parameter matrix. 

During training, only $\mathbf{v}_1, \mathbf{v}_2, \mathbf{v}_3$ and $\mathbf{W}$ are optimized. Therefore, compared with P-tuning and P-tuning v2, our method is  more efficient in terms of memory and time since the extra parameters do not participate in the calculations inside the PLM and the input length is also not increased. In Section~\ref{main—res}, we will provide a detailed analysis.

 \section{Experiments}
 \subsection{Setup}
  \label{setup}
 \paragraph{Datasets \& Baselines}
 The main experiments are conducted on diverse datasets to demonstrate that our discovery is general. For sentiment classification task, we use IMDB\footnote{\url{https://www.kaggle.com/lakshmi25npathi/sentiment-analysis-of-imdb-movie-reviews}} and SST~\cite{sst}. For paraphrase detection task, we use MRPC and QQP~\cite{wang-etal-2018-glue}. For natural language inference tasks, we use QNLI, RTE~\cite{wang-etal-2018-glue}, SciTaiL (SCITA)~\cite{Khot2018SciTaiLAT}, CB, BOOLQ ~\cite{superglue}. 
 The details of these datasets are listed in Table~\ref{tab:dataset}. 
 We compare our method with P-tuning~\cite{power_prompt} and P-tuning v2~\cite{liu2021ptuning}. Due to the unavailability of source code and for the sake of fair comparison, we reimplement these two models. We also report results obtained by normal fine tuning and Adapter tuning~\cite{add7}~\footnote{For adapter tuning, we use the library provided by ~\citet{add6}.}. Moreover, the baseline \textbf{Avg} which uses the average of the final layer's hidden states to conduct classification, is also added. 
For other tasks, we use accuracy as the evaluation metric.
 
 \begin{table}[t]
  \centering
    \resizebox{1\linewidth}{!}{\begin{tabular}{c|c|c|c|c|c}
    \hline 
    task&dataset&domain&\#train&\#validation&\#test \\
    \hline 
    sentiment
    &IMDB&movie review&45000&-&5000 \\
    classification&SST&movie review&67350&-&873 \\ 
    \hline 
    paraphrase
    &MRPC&news&3068&-&1721 \\
    detection &QQP&quora question&363868&-&40432 \\
    
    \hline 
    \multirow{5}{*}{\makecell{natural \\ language\\inference}}
    &QNLI&from SQuAD&105000&-&5000 \\
    &RTE&news, wiki&2490&-&277 \\
    &SciTaiL&sci exam, wiki&23596&-&2126 \\
    &CB&various&250&-&56 \\
    &BOOLQ&google query, wiki&94276&-&3270 \\
    \hline 
    
    \hline  
    \end{tabular}}
    \caption{Statistics of Datasets. \#train represents the number of training samples. Since the target of test data is unavailable, we use validation data as test data.}
    \label{tab:dataset}
\end{table}
 
 \paragraph{Implementation Details}
We conduct experiments on four different language models, i.e., RoBERTa-\{base, large\}, BERT-\{base, large\}. Each input is trimmed with a maximum length 160. The training batch size is 60. We  determine the number of training epochs according to the size of the training set since we find that low-resource datasets demand much more epochs to achieve a relatively better performance. The training epoch for IMDB, SST, SciTaiL is 50, for QNLI, QQP is 10. For other tasks, the training epoch is 100. The prompt length for P-tuning and P-tuning is 50 and the prompt vectors are randomly initialised. The learning rate for all the models are set to 0.01. For fine tuning, the learning rate is set to 1e-5. Generally, P-tuning, P-tuning v2 and our approach need larger learning rate to converge while fine tuning needs smaller learning rate. We use Adam optimizer~\cite{Adam}.
We run all the model using a single NVIDIA V100 GPU.
\subsection{Results \& Analysis}
\label{main—res}
 \begin{table*}[t]
 \setlength{\abovecaptionskip}{1pt}
    \large 
    \centering
    \resizebox{\linewidth}{!}{\begin{tabular}{c|ccccccccc|c}
        \hline 
        \hline 
        \multicolumn{10}{c}{RoBERTa-large} \\
        \hline
        \textbf{Model}& \textbf{IMDB}&\textbf{SST}&\textbf{MRPC}&\textbf{QQP}&\textbf{QNLI}&\textbf{SciTaiL}&\textbf{RTE}&\textbf{BOOLQ}&\textbf{CB}&AVERAGE\\
        \hline 
        P-tuning~\cite{power_prompt}& \textbf{94.34}& 94.72& 68.56& 88.43& 89.82
& 94.63& 61.01& 64.06& 87.50 &82.56\\
        P-tuning v2~\cite{liu2021ptuning}& 93.86& 95.29& 88.55& 87.13& 92.06& 95.71& 85.55& 74.77& 91.07&89.33\\
        Ours& 93.86& 94.72& 84.95& 86.22& 87.12& 91.90& 79.78& 75.59& 92.85&87.44\\
        Fine tuning& 94.12& \textbf{96.33}& 87.79& \textbf{92.09}& \textbf{93.77}& 96.04& 84.11& 80.42& 91.07 &90.63\\
        Avg& 85.54& 85.09& 68.97&76.81 &67.22& 73.75& 61.01&62.17&69.64&72.24\\
        Adapter~\cite{add7}&94.03&96.33& \textbf{89.95}&89.96&92.31& \textbf{96.54}& \textbf{86.64}& \textbf{84.06}& \textbf{94.64}&\colorbox{red!30}{91.60}\\
        \hline
        \hline 
        \multicolumn{10}{c}{RoBERTa-base} \\
        \hline 
        P-tuning~\cite{power_prompt}&92.20 & 93.69& 78.61& 87.77& 89.74& 89.93& 59.92& 71.16& 76.78&82.2\\
        P-tuning v2~\cite{liu2021ptuning}& \textbf{92.3}& 92.77& 83.43& 84.37& 89.53& 92.23& 73.28& 73.57& 85.71&85.24\\
       Ours& 90.38& 92.88 & 83.53& 84.65& 85.77& 88.71& 69.31&71.77 & 87.5&83.83\\
        Fine tuning& 92.12& \textbf{95.06}& \textbf{87.57}& \textbf{91.63}& \textbf{92.72}& 93.36& \textbf{75.09}& \textbf{79.17}& 89.28&\colorbox{red!30}{88.44}\\
         Avg& 83.12& 84.51& 67.98&75.89 &70.90& 69.23& 58.84&62.17&73.21&71.76\\
        Adapter~\cite{add7}&90.3& 94.38& 82.84&88.12& 90.52& \textbf{94.17}& 70.04& 76.51& \textbf{89.29}&86.24\\
        \hline 
        \hline 
        
        \multicolumn{10}{c}{BERT-large} \\
        \hline 
        P-tuning~\cite{power_prompt}& 90.84& 91.85& 70.48& 85.81& 85.16& 90.12 & 58.84& 66.17& 80.35&79.95\\
        P-tuning v2~\cite{liu2021ptuning}& \textbf{91.39}& 90.04& \textbf{84.36}& 85.45& 90.08& 91.81& \textbf{72.92}& 72.07& 85.71&84.87\\
       Ours& 90.92& 92.20& 82.27& 83.71& 87.12& 86.73& 72.20& 71.59& \textbf{92.85}&84.40\\
        Fine tuning&88.82& \textbf{93.92}& 83.84& \textbf{91.38}& \textbf{91.64}& 92.56& 71.11& \textbf{72.23}& 85.71&\colorbox{red!30}{85.69}\\
        Avg&84.04& 86.46& 69.84& 77.69& 67.29 & 75.72& 57.76& 63.48& 75.00&73.01\\
        Adapter~\cite{add7}&90.44& 93.57 & 83.09& 88.48& 89.32& \textbf{93.48}& 67.15& 70.76& 83.93&84.47\\
        \hline 
        \hline 
        
        \multicolumn{10}{c}{BERT-base} \\
        \hline 
        P-tuning~\cite{power_prompt}&88.58& 89.90& 74.14& 85.03& 87.56& 88.85& 63.89& 67.61& 75.00&80.06\\
        P-tuning v2~\cite{liu2021ptuning}& \textbf{89.58}& 90.71& 82.97& 87.08& 88.89& 90.02 & \textbf{70.03}& \textbf{71.19}& 82.14&\colorbox{red!30}{83.62}\\
       Ours& 88.62& 89.67& 80.82& 81.68& 84.86& 87.91& 66.78& 70.21& \textbf{92.85}&82.6\\
        Fine tuning&88.28& 88.41& 82.56& \textbf{90.83}& 84.39& 87.77& 67.14& 68.80& 91.07&83.25\\
    Avg &81.48& 84.51& 71.52& 76.35& 71.55& 77.65& 61.37& 63.21&71.42&73.22 \\
    Adapter~\cite{add7}&88.82& \textbf{92.32}& \textbf{84.07}& 87.49& \textbf{90.66}& \textbf{92.18}& 65.70& 68.72& 80.36&83.37\\
        \hline 
        \hline 
    \end{tabular}}
    \caption{Main Results. Best accuracy is boldfaced.}
    \label{tab:auto1}
\end{table*}
\paragraph{Main Results} The main results of baselines and ours are presented in Table~\ref{tab:auto1}.  
We firstly observe that fine tuning and adapter tuning can generally obtain better AVERAGE scores and Avg exhibits the worst performance. Since fine tuning tunes the whole model and adapter needs to tune  much more introduced parameters, the better performance is expected.
Next we analyse the performance of Ours and P-tuning (v2). Generally, Ours achieves higher AVERAGE score compared with P-tuning and slightly lower (within 2\%) AVERAGE score compared with P-tuning v2.  Specifically, Compared with P-tuning, Ours delivers much better performance on MRPC, RTE, BOOLQ, CB on all four language models. On IMDB and SST, Ours and P-tuning show similar performance. Relatively, the accuracy of Ours lags behind P-tuning on QQP, QNLI, SciTaiL on most language models. Compared with P-tuning v2, Ours achieves higher performance on CB and similar performance on IMDB, SST, MRPC, BOOLQ.  
The above results show that the raw hidden states have already contained most of information that is needed for classification. Although the performance of Ours is acceptable, it can not surpass P-tuning (v2) on QQP, QNLI, SciTaiL, which all need the model to predict the relationship of two sentences. The reason may be that the purification process of Ours does not have the ability of \textbf{reasoning} which is required by these datasets. For instance, at self-attention stage, we use a vector $\mathbf{v}_3$ to attend a sequence of word representations and this limits the ability of the model to make more complex reasoning. P-tuning (v2) can alleviate this problem by performing $(\mathbf{Q},\mathbf{K},\mathbf{V})$-format self-attention among prompt vectors and hidden states. Recall that more complex manipulations over raw hidden states (means more parameters may be introduced) can be performed, the performance may also be improved. We leave it as a future work.  

\begin{figure}
    \centering
    \includegraphics[width=0.8\linewidth]{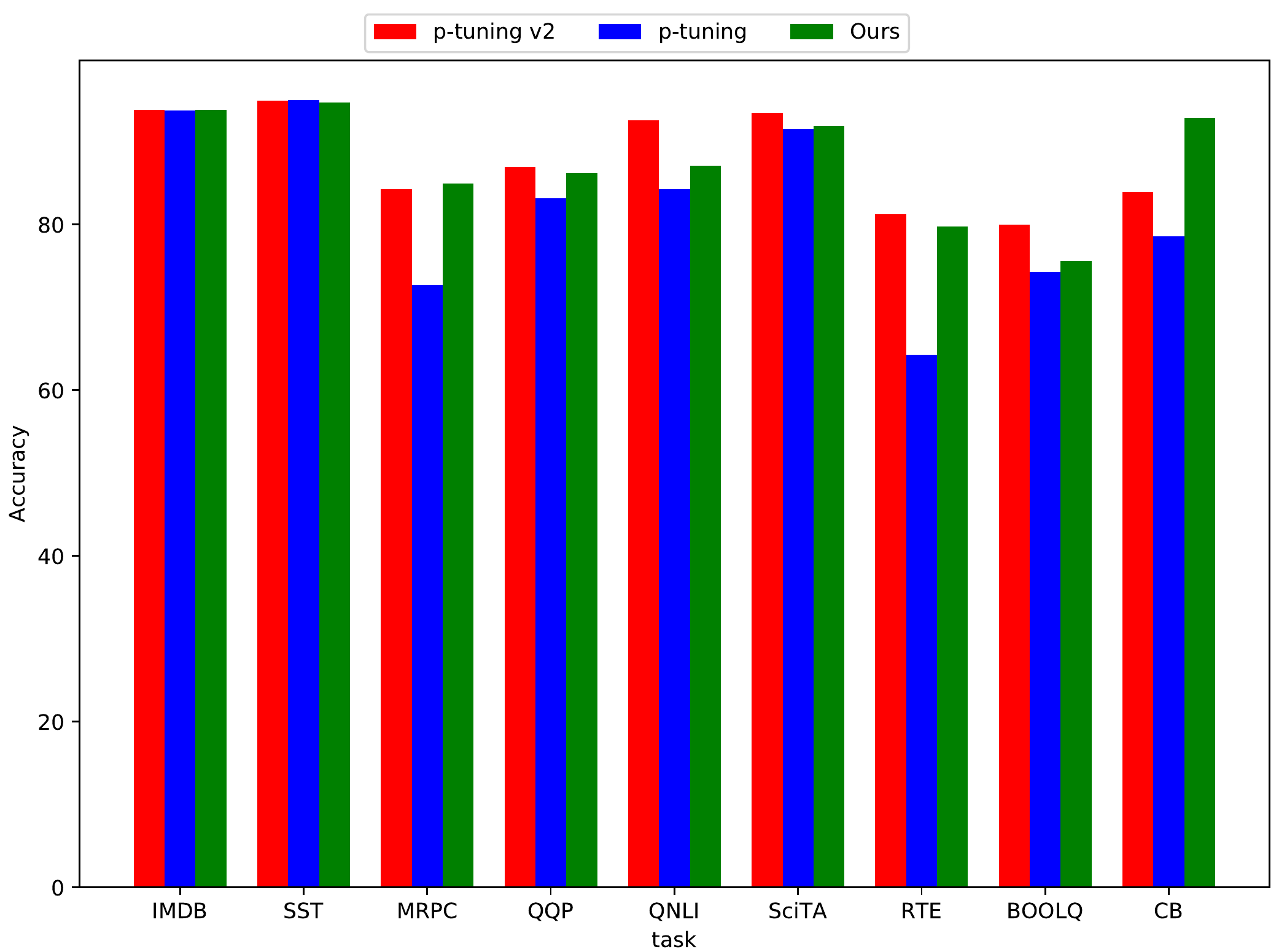}
    \caption{Results with Small Prompt Length}%
    \label{fig:samll-length}
\end{figure}

To examine the performance of these tuning methods with  approximate number of extra parameters, we decrease the number of prompt vectors in P-tuning to 3 and in P-tuning v2 to 1. The results are shown in Figure~\ref{fig:samll-length}. We can find that Ours is better than P-tuning on the whole and is comparable with P-tuning v2. We also attempt to increase the number of parameters of Ours, i.e., introducing multiple $\mathbf{v}_3$ to form multi-heads self-attention, but find no improvement of the performance. This may be one drawback of our method.  

\paragraph{Parameter \& Time Efficiency}
Although only three vectors are used in Ours, the performance is not bad. The number of parameters Ours used is $L+1+2d$ where $L+1$ is the numbers of transformer layers plus the word embedding layer. $d$ is the dimension of hidden states. The number of parameters used by P-tuning and P-tuning v2 are $Kd$ and $(L+1)Kd$ respectively where $K$ is the number of prompt vectors in each layer.  Moreover Ours is advantageous over P-tuning (v2) in terms of time and memory efficiency as shown in Table~\ref{tab:efficiency}. Compared with Ours, other tuning methods spend 2.5 times longer in training. For GPU usage, fine tuning uses 4 times memory while other two baselines use 2.5 times memory compared with Ours. 
There are two main reasons that can account to such results: \textbf{(i)} operations outside PLMs avoid the calculation and propagation of gradients inside the PLMs. \textbf{(ii)} our method does not increase the length of input which is an important factor for training time~\cite{geng2021is}.   
\begin{table}[t]
  \centering
    \resizebox{1\linewidth}{!}{\begin{tabular}{c|c|c|c|c}
    \hline 
    &Fine tuning&P-tuning&P-tuning v2&Ours \\
    \hline 
    paramater&- &$Kd$& $(L+1)Kd$&$L+1+2d$\\
    \hline 
    training time&2.50x& 2.50x&2.50x&1.00x \\
    \hline 
    inferring time& 1.00x& 1.01x & 1.03x &1.004x \\                
    \hline 
    GPU usage (training)&4.00x &2.50x &2.50x&1.00x\\ 
    \hline 
    \end{tabular}}
    \caption{Comparison of Time and Memory Cost. The prompt length for P-tuning and P-tuning v2 is 3 and 1 respectively.}
    \label{tab:efficiency}
\end{table}

  \begin{table*}[t]
  \centering
    \resizebox{0.8\linewidth}{!}{\begin{tabular}{c|c|c}
    \hline 
    \hline
    \multirow{3}{*}{SST}& i \textcolor{c4}{had} to \textcolor{c3}{look} \textcolor{c1}{away} - this was \textcolor{c5}{god} \textcolor{c2}{awful}.& 0\\
    &it does \textcolor{c2}{n't} believe in itself , it has \textcolor{c3}{no} \textcolor{c4}{sense} of humor\textcolor{c5}{...} it 's just plain \textcolor{c1}{bored}. & 0\\  
    &we know the plot 's a little crazy , \textcolor{c2}{but} \textcolor{c4}{it} \textcolor{c1}{held} my \textcolor{c3}{interest} \textcolor{c5}{from} start to finish. &1 \\
    \hline 
    \multirow{2}{*}{SCITA}&\textcolor{c2}{Front}- The \textcolor{c1}{boundary} between two different \textcolor{c3}{air} masses. & \multirow{2}{*}{1}\\
    & In \textcolor{c3}{weather} terms, the boundary between two air masses \textcolor{c2}{is} called \textcolor{c1}{front}. \\
    \hline 
    \multirow{2}{*}{QQP}&\textcolor{c2}{Who} is the \textcolor{c4}{best} graphic designer \textcolor{c3}{in} \textcolor{c1}{israel}? & \multirow{2}{*}{0} \\
    &What \textcolor{c2}{is} \textcolor{c3}{the} best \textcolor{c1}{forum} for graphic \textcolor{c3}{designers}? \\ 
    \hline 
    \hline 
    \end{tabular}}
    \caption{Visualization of Self-Attention.}
    \label{tab:vis_attn}
\end{table*}

\subsection{Ablation Study}
\begin{figure}[t]
 \centering
  \includegraphics[width=0.8\linewidth]{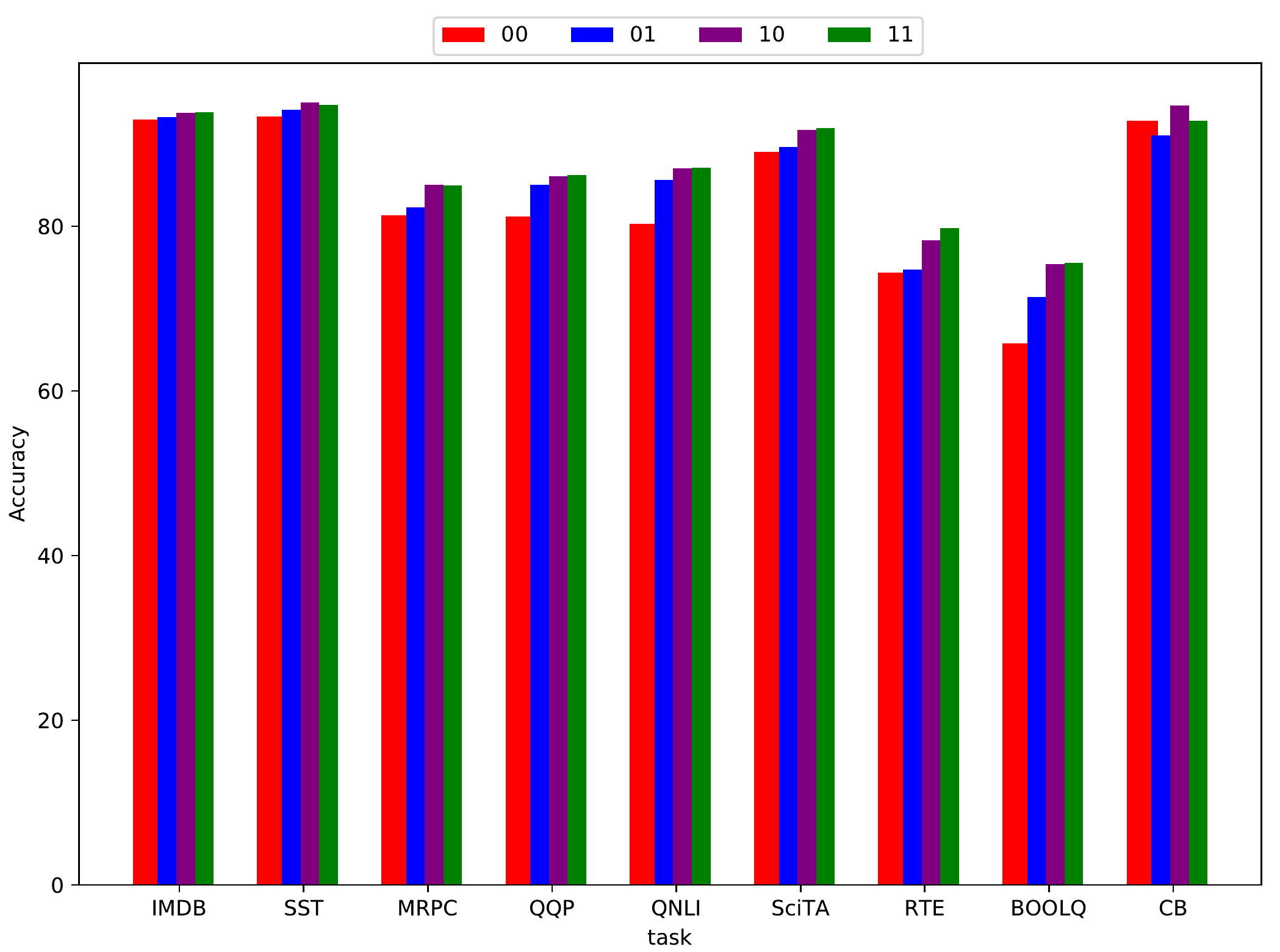}
  \caption{Importance of Parameter $\mathbf{v}_1$ and $\mathbf{v}_2$. The first digit and the second digit (from left to right) represents whether $\mathbf{v}_1$ and $\mathbf{v}_2$ are introduced.}
  \label{fig:aba-1}
\end{figure}

\paragraph{Necessity of $\mathbf{v}_1$ and $\mathbf{v}_2$} We investigate the importance of parameter $\mathbf{v}_1$ and $\mathbf{v}_2$. The result is shown in Figure~\ref{fig:aba-1}. We can see that generally better performance can be achieved when $\mathbf{v}_1$ and $\mathbf{v}_2$ are both introduced. Compared with ``00'', the performance is improved when $\mathbf{v}_2$ is introduced (``01''). When $\mathbf{v}_1$ is introduced, the improvement is not so significant when $\mathbf{v}_2$ is added. The reason may be that the parameter $\mathbf{W}$ of the classification head can also learn to assign a higher value if a position in hidden states is important for this task. We tend to add $\mathbf{v}_2$ to achieve a slightly better performance. 

\begin{figure}[t]
    \centering
    
    \includegraphics[width=0.8\linewidth]{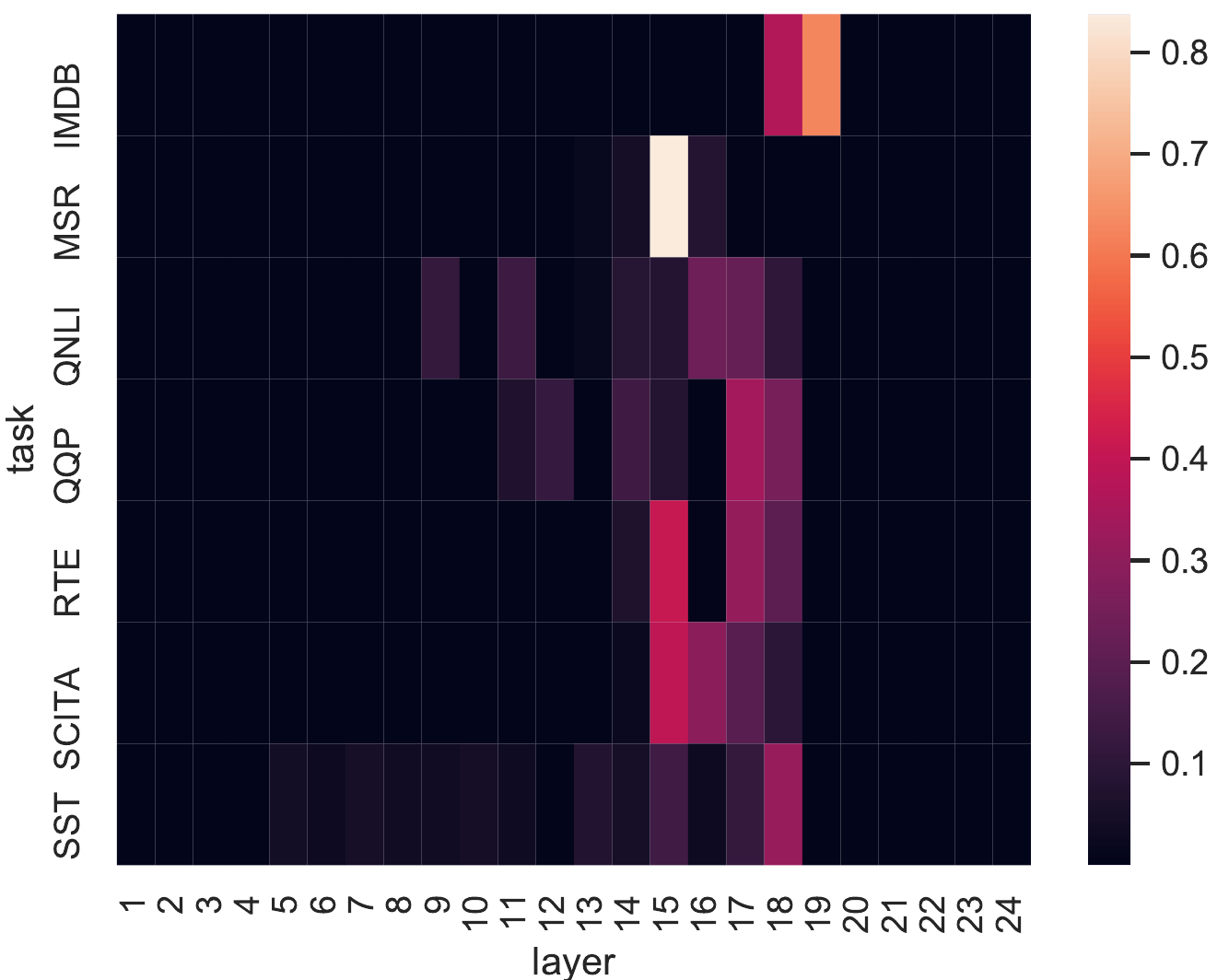}
    \caption{Visualization of Softmax-normalised $\mathbf{v}_1$}%
    \label{fig:vis-attn}
\end{figure}

\begin{figure}[t]
    \centering
    
    \subfloat[\centering Full-Size (1024)]{{\includegraphics[width=0.42\linewidth]{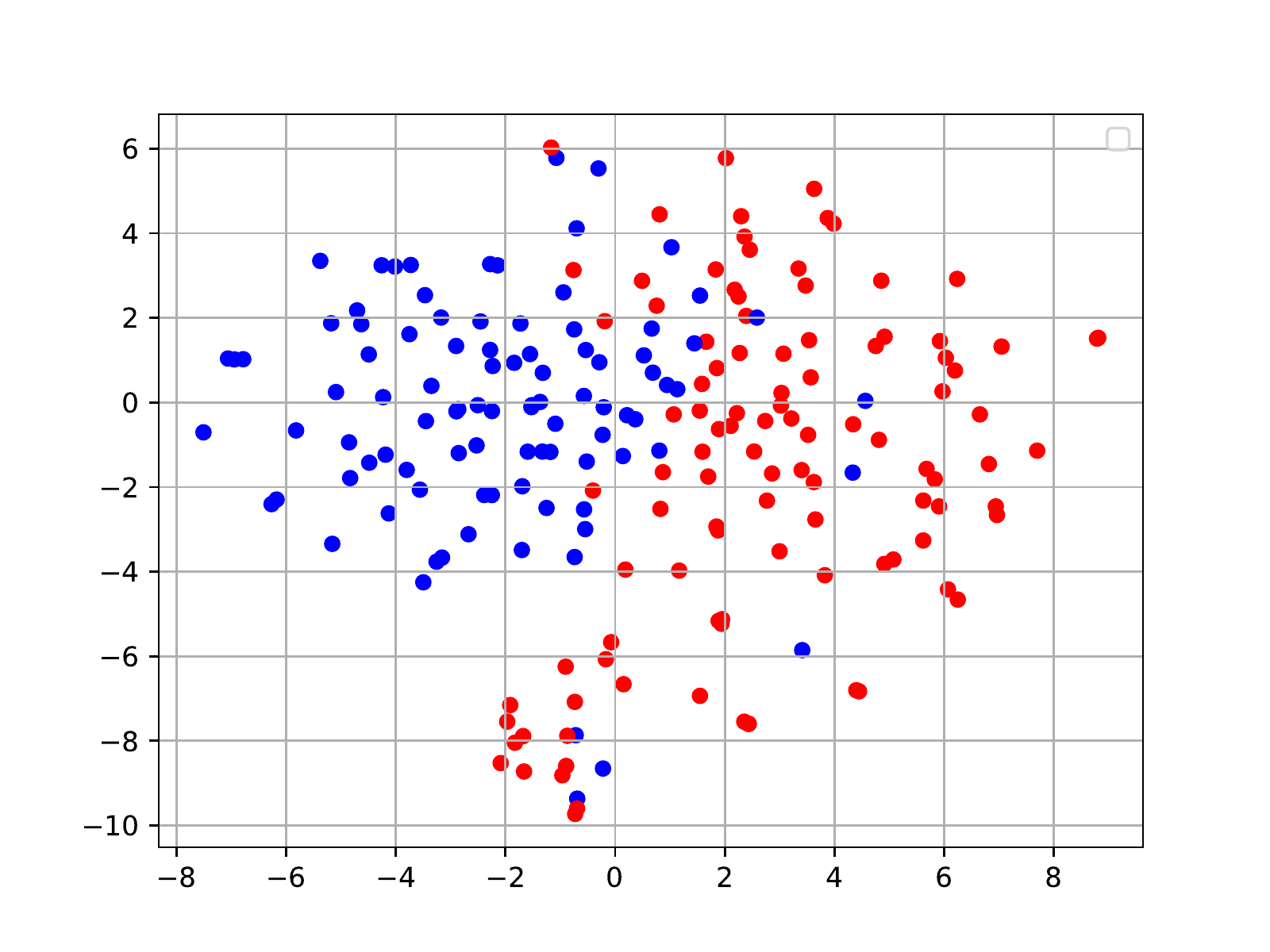} }}
    \qquad
    \subfloat[\centering  Top-100]{{\includegraphics[width=0.42\linewidth]{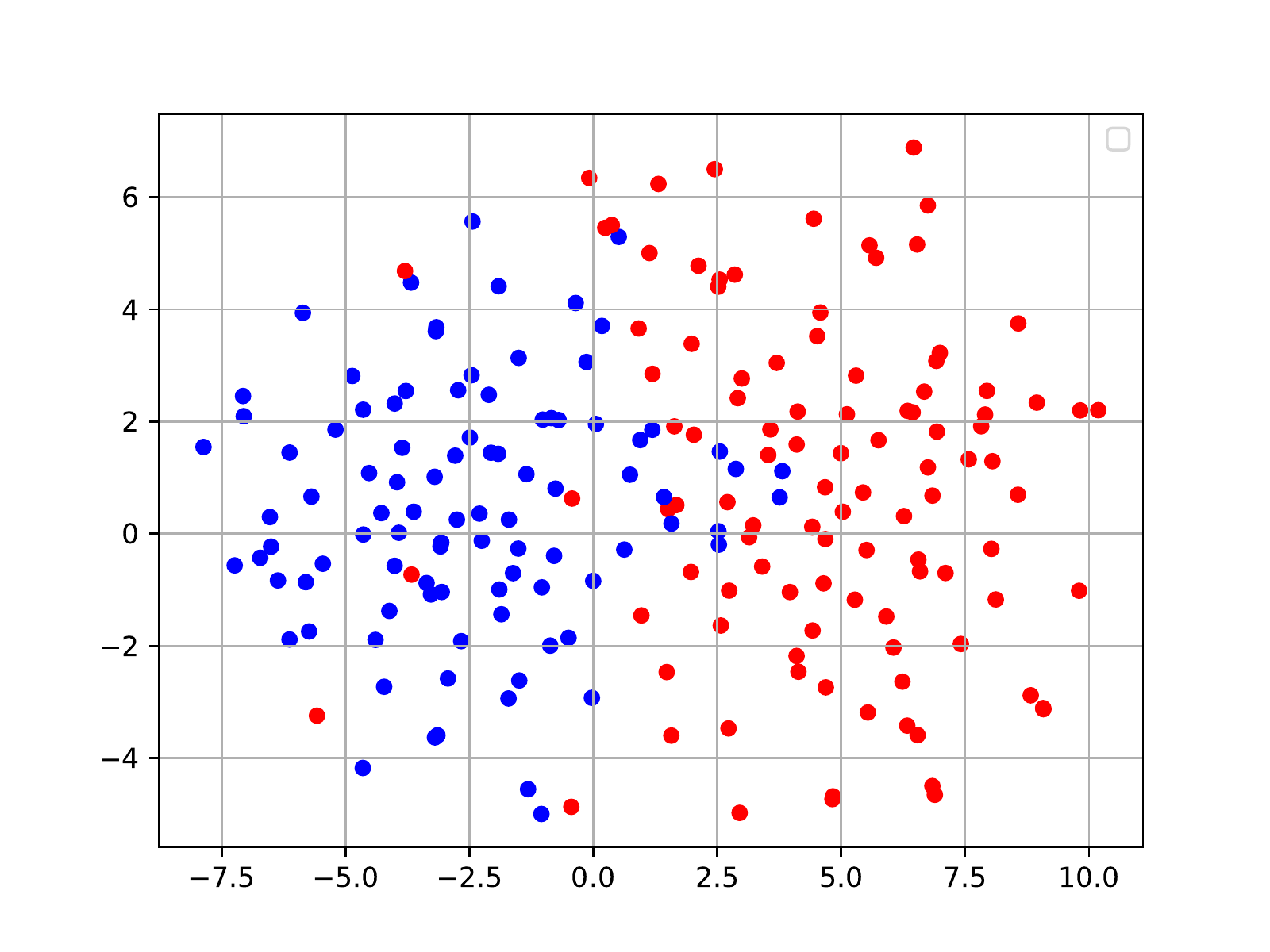} }}
    \caption{t-SNE Visualization of Hidden States. We present the full-size hidden states visualization and Top-100 (selected according to the value in $\mathbf{v}_2$) digits of hidden states. Blue and red represent two classes.}%
    \label{fig:vis-subspace}
\end{figure}

\paragraph{Layer Weight Visualization}
Different layers in PLMs exert different impact on different tasks~\cite{add3, add4}.
We inspect which layer is most important for each task and the result is illustrated in Figure~\ref{fig:vis-attn}. We can see that most tasks require one or multiple layers located between 15 and 18, which are the middle layers. We speculate that this is because these tasks are neither simple nor complex. 
\paragraph{Subspace Visualization}
$\textbf{v}_2$ is used to conduct soft mask over hidden states since we argue that a subset of dimensions of hidden states represents a specific kind of information, e.g., sentiment. We visualise the subspace in Figure~\ref{fig:vis-subspace}. The hidden states are obtained based on the framework in Figure~\ref{fig:my_model} except that step  \textbf{\circled{\small 2}} is omitted. We can see that compared with (a), the red and glue points are farther with each other in (b), which shows that subspace is better than full space. 
\paragraph{Attention Visualization}
$\mathbf{v}_3$ is used to attend the most important part of the input. We list some examples to investigate what are learned by $\mathbf{v}_3$ in Table~\ref{tab:vis_attn}. A darker color means that the word has a larger weight. We can observe that $\mathbf{v}_3$ can generally attend the most important words that are critical for classification. But as we have stated above, the reasoning ability of this kind of self-attention is weak especially for tasks with paired sentences. 

\subsection{Transfer Ability}
  \begin{table}[t]
 \setlength{\abovecaptionskip}{1pt}
    \large 
    \centering
    \resizebox{\linewidth}{!}{\begin{tabular}{c|cccc}
        \hline 
        \hline
        \textbf{Tasks}&P-tuning&P-tuning v2&Fine tuning&Ours\\
        \hline 
        IMDB $\Rightarrow$ SST& 90.48& 89.44& 85.77& \textbf{90.48}\\
        SST $\Rightarrow$ IMDB& 84.16& \textbf{91.56}& 91.18& 88.1 \\
        QNLI$\Rightarrow$ SciTaiL& 54.79& 61.24& \textbf{76.38}& 70.13 \\
        SciTaiL$\Rightarrow$ QNLI& \textbf{56.17}& 52.41& 51.13& 54.06 \\
        SciTaiL$\Rightarrow$ RTE & 66.06& 68.95& \textbf{76.17}& 66.06\\
        \hline
        \hline 
    \end{tabular}}
    \captionsetup{justification=centering}
    \caption{Transfer Learning}
    \label{tab:transfer}
\end{table}
We investigate the transfer ability of tuning strategies under the same task type. We optimize the trainable parameters, including the parameters of the classification head, on the source task and evaluate the whole model on the target task. The implementation details are the same as that stated in Section~\ref{setup}. The language model we use is RoBERTa-large. The results are listed in Table~\ref{tab:transfer}. We can find that there is not a tuning method that can achieve the best performance on all transfer learning tasks. Ours and other tuning methods have a comparable performance.

\subsection{Few Shot Settings}
\begin{table}[t]
 \setlength{\abovecaptionskip}{1pt}
    \large 
    \centering
    \resizebox{\linewidth}{!}{\begin{tabular}{c|cccc}
        \hline 
        \hline
        \textbf{Tasks}&P-tuning&P-tuning v2&Fine tuning&Ours\\
        \hline 
        IMDB&  75.28& 81.80& \textbf{88.22} &86.76\\
        MSR& 66.47& 66.64& \textbf{71.12}&69.72\\
        SciTaiL& 68.01 & 69.09& \textbf{80.15}& 69.42\\
        \hline
        \hline 
    \end{tabular}}
    \captionsetup{justification=centering}
    \caption{Few Shot Scenario. 
    }
    \label{tab:few-shot}
\end{table}
We explore the performance of different tuning methods under few shot scenario. We select three tasks and for each task, we randomly select 64 samples as the training set and the evaluation set is kept original. The prompts length of P-tuning and P-tuning v2 is set to 3 and 1 respectively to alleviate overfitting issue. As before, we use RoBERTa-large as the language model. The results are shown in Table~\ref{tab:few-shot}. Generally, fine tuning outperforms the other parameter efficient tuning methods. ignoring fine tuning, we can observe that Ours surpasses P-tuning and P-tuning v2. 

\subsection{Privacy Protection}
Making a PLM public, i.e., the architecture and all the parameters in the PLM are known to us, often raises the privacy issue~\footnote{\url{https://ai.googleblog.com/2020/12/privacy-considerations-in-large.html}}. However, if the PLMs are not publicly accessible, it is difficult to (fine, prompt, adapter)-tune the model for downstream tasks since the back propagation can not be performed.
Our proposed method can be regarded as a black-box tuning method~\cite{sun2022blackbox}. Therefore it can be applied in a scenario where PLMs are designed as an API service. All we need to do is to feed the input into the API to obtain the hidden states and then train a task-specific classifier with these raw hidden states. Since individuals have no access to the internal parameters of the PLMs, privacy is well protected.   
\section{Conclusion}
We design a simple method that aggregates raw hidden states utilizing three vectors for each task and conduct extensive experiments on an assortment of tasks to show that the aggregated state(s) is(are) the information needed for classification. Our method can achieve comparable performance with P-tuning and P-tuning v2. This can partially explain why continuous prompt tuning methods can distill useful information from PLMs with only a few prompt vectors. Further work should aim to explore whether our finding can be applied to more complex tasks, e.g., QA. Moreover, methods of aggregation raw hidden states can also be upgraded to have more reasoning ability while introducing acceptable number of parameters.  
\bibliography{anthology,custom}

\begin{thebibliography}{31}
\expandafter\ifx\csname natexlab\endcsname\relax\def\natexlab#1{#1}\fi

\bibitem[{Biesialska et~al.(2020)Biesialska, Biesialska, and
  Costa-juss{\`a}}]{biesialska-etal-2020-continual}
Magdalena Biesialska, Katarzyna Biesialska, and Marta~R. Costa-juss{\`a}. 2020.
\newblock \href {https://doi.org/10.18653/v1/2020.coling-main.574} {Continual
  lifelong learning in natural language processing: A survey}.
\newblock In \emph{Proceedings of the 28th International Conference on
  Computational Linguistics}, pages 6523--6541, Barcelona, Spain (Online).
  International Committee on Computational Linguistics.

\bibitem[{Brown et~al.(2020)Brown, Mann, Ryder, Subbiah, Kaplan, Dhariwal,
  Neelakantan, Shyam, Sastry, Askell, Agarwal, Herbert-Voss, Krueger, Henighan,
  Child, Ramesh, Ziegler, Wu, Winter, Hesse, Chen, Sigler, Litwin, Gray, Chess,
  Clark, Berner, McCandlish, Radford, Sutskever, and
  Amodei}]{NEURIPS2020_1457c0d6}
Tom Brown, Benjamin Mann, Nick Ryder, Melanie Subbiah, Jared~D Kaplan, Prafulla
  Dhariwal, Arvind Neelakantan, Pranav Shyam, Girish Sastry, Amanda Askell,
  Sandhini Agarwal, Ariel Herbert-Voss, Gretchen Krueger, Tom Henighan, Rewon
  Child, Aditya Ramesh, Daniel Ziegler, Jeffrey Wu, Clemens Winter, Chris
  Hesse, Mark Chen, Eric Sigler, Mateusz Litwin, Scott Gray, Benjamin Chess,
  Jack Clark, Christopher Berner, Sam McCandlish, Alec Radford, Ilya Sutskever,
  and Dario Amodei. 2020.
\newblock \href
  {https://proceedings.neurips.cc/paper/2020/file/1457c0d6bfcb4967418bfb8ac142f64a-Paper.pdf}
  {Language models are few-shot learners}.
\newblock In \emph{Advances in Neural Information Processing Systems},
  volume~33, pages 1877--1901. Curran Associates, Inc.

\bibitem[{Dathathri et~al.(2020)Dathathri, Madotto, Lan, Hung, Frank, Molino,
  Yosinski, and Liu}]{add2}
Sumanth Dathathri, Andrea Madotto, Janice Lan, Jane Hung, Eric Frank, Piero
  Molino, Jason Yosinski, and Rosanne Liu. 2020.
\newblock \href {https://openreview.net/forum?id=H1edEyBKDS} {Plug and play
  language models: A simple approach to controlled text generation}.
\newblock In \emph{International Conference on Learning Representations}.

\bibitem[{Devlin et~al.(2019)Devlin, Chang, Lee, and
  Toutanova}]{devlin-etal-2019-bert}
Jacob Devlin, Ming-Wei Chang, Kenton Lee, and Kristina Toutanova. 2019.
\newblock \href {https://doi.org/10.18653/v1/N19-1423} {{BERT}: Pre-training of
  deep bidirectional transformers for language understanding}.
\newblock In \emph{Proceedings of the 2019 Conference of the North {A}merican
  Chapter of the Association for Computational Linguistics: Human Language
  Technologies, Volume 1 (Long and Short Papers)}, pages 4171--4186,
  Minneapolis, Minnesota. Association for Computational Linguistics.

\bibitem[{Ethayarajh(2019)}]{add4}
Kawin Ethayarajh. 2019.
\newblock \href {https://doi.org/10.18653/v1/D19-1006} {How contextual are
  contextualized word representations? {C}omparing the geometry of {BERT},
  {ELM}o, and {GPT}-2 embeddings}.
\newblock In \emph{Proceedings of the 2019 Conference on Empirical Methods in
  Natural Language Processing and the 9th International Joint Conference on
  Natural Language Processing (EMNLP-IJCNLP)}, pages 55--65, Hong Kong, China.
  Association for Computational Linguistics.

\bibitem[{Geng et~al.(2021)Geng, Guo, Chen, Li, Wei, and Lin}]{geng2021is}
Zhengyang Geng, Meng-Hao Guo, Hongxu Chen, Xia Li, Ke~Wei, and Zhouchen Lin.
  2021.
\newblock \href {https://openreview.net/forum?id=1FvkSpWosOl} {Is attention
  better than matrix decomposition?}
\newblock In \emph{International Conference on Learning Representations}.

\bibitem[{Hochreiter and Schmidhuber(1997)}]{HochSchm97}
Sepp Hochreiter and Jürgen Schmidhuber. 1997.
\newblock Long short-term memory.
\newblock \emph{Neural Computation}, 9(8):1735--1780.

\bibitem[{Jawahar et~al.(2019)Jawahar, Sagot, and
  Seddah}]{jawahar-etal-2019-bert}
Ganesh Jawahar, Beno{\^\i}t Sagot, and Djam{\'e} Seddah. 2019.
\newblock \href {https://doi.org/10.18653/v1/P19-1356} {What does {BERT} learn
  about the structure of language?}
\newblock In \emph{Proceedings of the 57th Annual Meeting of the Association
  for Computational Linguistics}, pages 3651--3657, Florence, Italy.
  Association for Computational Linguistics.

\bibitem[{Khot et~al.(2018)Khot, Sabharwal, and Clark}]{Khot2018SciTaiLAT}
Tushar Khot, Ashish Sabharwal, and Peter Clark. 2018.
\newblock Scitail: A textual entailment dataset from science question
  answering.
\newblock In \emph{AAAI}.

\bibitem[{Kingma and Ba(2014)}]{Adam}
Diederik~P. Kingma and Jimmy Ba. 2014.
\newblock \href {http://arxiv.org/abs/1412.6980} {Adam: A method for stochastic
  optimization}.
\newblock Cite arxiv:1412.6980Comment: Published as a conference paper at the
  3rd International Conference for Learning Representations, San Diego, 2015.

\bibitem[{Lester et~al.(2021)Lester, Al{-}Rfou, and Constant}]{power_prompt}
Brian Lester, Rami Al{-}Rfou, and Noah Constant. 2021.
\newblock \href {http://arxiv.org/abs/2104.08691} {The power of scale for
  parameter-efficient prompt tuning}.
\newblock \emph{CoRR}, abs/2104.08691.

\bibitem[{Li and Liang(2021)}]{li-liang-2021-prefix}
Xiang~Lisa Li and Percy Liang. 2021.
\newblock \href {https://doi.org/10.18653/v1/2021.acl-long.353} {Prefix-tuning:
  Optimizing continuous prompts for generation}.
\newblock In \emph{Proceedings of the 59th Annual Meeting of the Association
  for Computational Linguistics and the 11th International Joint Conference on
  Natural Language Processing (Volume 1: Long Papers)}, Online. Association for
  Computational Linguistics.

\bibitem[{Lin et~al.(2019)Lin, Tan, and Frank}]{add3}
Yongjie Lin, Yi~Chern Tan, and Robert Frank. 2019.
\newblock \href {https://doi.org/10.18653/v1/W19-4825} {Open sesame: Getting
  inside {BERT}{'}s linguistic knowledge}.
\newblock In \emph{Proceedings of the 2019 ACL Workshop BlackboxNLP: Analyzing
  and Interpreting Neural Networks for NLP}, pages 241--253, Florence, Italy.
  Association for Computational Linguistics.

\bibitem[{Liu et~al.(2021{\natexlab{a}})Liu, Yuan, Fu, Jiang, Hayashi, and
  Neubig}]{liu_pretrain_prompt}
Pengfei Liu, Weizhe Yuan, Jinlan Fu, Zhengbao Jiang, Hiroaki Hayashi, and
  Graham Neubig. 2021{\natexlab{a}}.
\newblock \href {http://arxiv.org/abs/2107.13586} {Pre-train, prompt, and
  predict: {A} systematic survey of prompting methods in natural language
  processing}.
\newblock \emph{CoRR}, abs/2107.13586.

\bibitem[{Liu et~al.(2021{\natexlab{b}})Liu, Ji, Fu, Du, Yang, and
  Tang}]{liu2021ptuning}
Xiao Liu, Kaixuan Ji, Yicheng Fu, Zhengxiao Du, Zhilin Yang, and Jie Tang.
  2021{\natexlab{b}}.
\newblock \href {http://arxiv.org/abs/2110.07602} {P-tuning v2: Prompt tuning
  can be comparable to fine-tuning universally across scales and tasks}.

\bibitem[{Liu et~al.(2019{\natexlab{a}})Liu, He, Chen, and Gao}]{liu2019mt-dnn}
Xiaodong Liu, Pengcheng He, Weizhu Chen, and Jianfeng Gao. 2019{\natexlab{a}}.
\newblock \href {https://www.aclweb.org/anthology/P19-1441} {Multi-task deep
  neural networks for natural language understanding}.
\newblock In \emph{Proceedings of the 57th Annual Meeting of the Association
  for Computational Linguistics}, pages 4487--4496, Florence, Italy.
  Association for Computational Linguistics.

\bibitem[{Liu et~al.(2019{\natexlab{b}})Liu, Ott, Goyal, Du, Joshi, Chen, Levy,
  Lewis, Zettlemoyer, and Stoyanov}]{liu2019roberta}
Yinhan Liu, Myle Ott, Naman Goyal, Jingfei Du, Mandar Joshi, Danqi Chen, Omer
  Levy, Mike Lewis, Luke Zettlemoyer, and Veselin Stoyanov. 2019{\natexlab{b}}.
\newblock Roberta: A robustly optimized bert pretraining approach.
\newblock \emph{arXiv preprint arXiv:1907.11692}.

\bibitem[{Peters et~al.(2018)Peters, Neumann, Iyyer, Gardner, Clark, Lee, and
  Zettlemoyer}]{peters-etal-2018-deep}
Matthew~E. Peters, Mark Neumann, Mohit Iyyer, Matt Gardner, Christopher Clark,
  Kenton Lee, and Luke Zettlemoyer. 2018.
\newblock \href {https://doi.org/10.18653/v1/N18-1202} {Deep contextualized
  word representations}.
\newblock In \emph{Proceedings of the 2018 Conference of the North {A}merican
  Chapter of the Association for Computational Linguistics: Human Language
  Technologies, Volume 1 (Long Papers)}, pages 2227--2237, New Orleans,
  Louisiana. Association for Computational Linguistics.

\bibitem[{Petroni et~al.(2019)Petroni, Rockt{\"a}schel, Riedel, Lewis, Bakhtin,
  Wu, and Miller}]{petroni-etal-2019-language}
Fabio Petroni, Tim Rockt{\"a}schel, Sebastian Riedel, Patrick Lewis, Anton
  Bakhtin, Yuxiang Wu, and Alexander Miller. 2019.
\newblock \href {https://doi.org/10.18653/v1/D19-1250} {Language models as
  knowledge bases?}
\newblock In \emph{Proceedings of the 2019 Conference on Empirical Methods in
  Natural Language Processing and the 9th International Joint Conference on
  Natural Language Processing (EMNLP-IJCNLP)}, pages 2463--2473, Hong Kong,
  China. Association for Computational Linguistics.

\bibitem[{Pfeiffer et~al.(2020{\natexlab{a}})Pfeiffer, R{\"u}ckl{\'e}, Poth,
  Kamath, Vuli{\'c}, Ruder, Cho, and Gurevych}]{add6}
Jonas Pfeiffer, Andreas R{\"u}ckl{\'e}, Clifton Poth, Aishwarya Kamath, Ivan
  Vuli{\'c}, Sebastian Ruder, Kyunghyun Cho, and Iryna Gurevych.
  2020{\natexlab{a}}.
\newblock Adapterhub: A framework for adapting transformers.
\newblock In \emph{Proceedings of the 2020 Conference on Empirical Methods in
  Natural Language Processing: System Demonstrations}, pages 46--54.

\bibitem[{Pfeiffer et~al.(2020{\natexlab{b}})Pfeiffer, Vuli{\'c}, Gurevych, and
  Ruder}]{add7}
Jonas Pfeiffer, Ivan Vuli{\'c}, Iryna Gurevych, and Sebastian Ruder.
  2020{\natexlab{b}}.
\newblock \href {https://doi.org/10.18653/v1/2020.emnlp-main.617} {{MAD-X}:
  {A}n {A}dapter-{B}ased {F}ramework for {M}ulti-{T}ask {C}ross-{L}ingual
  {T}ransfer}.
\newblock In \emph{Proceedings of the 2020 Conference on Empirical Methods in
  Natural Language Processing (EMNLP)}, pages 7654--7673, Online. Association
  for Computational Linguistics.

\bibitem[{Qiu et~al.(2020)Qiu, Sun, Xu, Shao, Dai, and Huang}]{PLM_survey}
Xipeng Qiu, Tianxiang Sun, Yige Xu, Yunfan Shao, Ning Dai, and Xuanjing Huang.
  2020.
\newblock \href {http://arxiv.org/abs/2003.08271} {Pre-trained models for
  natural language processing: {A} survey}.
\newblock \emph{CoRR}, abs/2003.08271.

\bibitem[{Raffel et~al.(2020)Raffel, Shazeer, Roberts, Lee, Narang, Matena,
  Zhou, Li, and Liu}]{2020t5}
Colin Raffel, Noam Shazeer, Adam Roberts, Katherine Lee, Sharan Narang, Michael
  Matena, Yanqi Zhou, Wei Li, and Peter~J. Liu. 2020.
\newblock \href {http://jmlr.org/papers/v21/20-074.html} {Exploring the limits
  of transfer learning with a unified text-to-text transformer}.
\newblock \emph{Journal of Machine Learning Research}, 21(140):1--67.

\bibitem[{Schick and Sch{\"u}tze(2021)}]{schick-schutze-2021-exploiting}
Timo Schick and Hinrich Sch{\"u}tze. 2021.
\newblock \href {https://aclanthology.org/2021.eacl-main.20} {Exploiting
  cloze-questions for few-shot text classification and natural language
  inference}.
\newblock In \emph{Proceedings of the 16th Conference of the European Chapter
  of the Association for Computational Linguistics: Main Volume}, pages
  255--269, Online. Association for Computational Linguistics.

\bibitem[{Socher et~al.(2013)Socher, Perelygin, Wu, Chuang, Manning, Ng, and
  Potts}]{sst}
Richard Socher, Alex Perelygin, Jean Wu, Jason Chuang, Christopher~D. Manning,
  Andrew Ng, and Christopher Potts. 2013.
\newblock \href {https://aclanthology.org/D13-1170} {Recursive deep models for
  semantic compositionality over a sentiment treebank}.
\newblock In \emph{Proceedings of the 2013 Conference on Empirical Methods in
  Natural Language Processing}, pages 1631--1642, Seattle, Washington, USA.
  Association for Computational Linguistics.

\bibitem[{Subramani et~al.(2019)Subramani, Bowman, and Cho}]{add1}
Nishant Subramani, Samuel~R. Bowman, and Kyunghyun Cho. 2019.
\newblock \href {http://arxiv.org/abs/1907.04944} {Can unconditional language
  models recover arbitrary sentences?}
\newblock \emph{CoRR}, abs/1907.04944.

\bibitem[{Sun et~al.(2022)Sun, Shao, Qian, Huang, and Qiu}]{sun2022blackbox}
Tianxiang Sun, Yunfan Shao, Hong Qian, Xuanjing Huang, and Xipeng Qiu. 2022.
\newblock \href {http://arxiv.org/abs/2201.03514} {Black-box tuning for
  language-model-as-a-service}.

\bibitem[{Vaswani et~al.(2017)Vaswani, Shazeer, Parmar, Uszkoreit, Jones,
  Gomez, Kaiser, and Polosukhin}]{attnneed}
Ashish Vaswani, Noam Shazeer, Niki Parmar, Jakob Uszkoreit, Llion Jones,
  Aidan~N. Gomez, \L{}ukasz Kaiser, and Illia Polosukhin. 2017.
\newblock Attention is all you need.
\newblock In \emph{Proceedings of the 31st International Conference on Neural
  Information Processing Systems}, NIPS'17, page 6000–6010, Red Hook, NY,
  USA. Curran Associates Inc.

\bibitem[{Wang et~al.(2019)Wang, Pruksachatkun, Nangia, Singh, Michael, Hill,
  Levy, and Bowman}]{superglue}
Alex Wang, Yada Pruksachatkun, Nikita Nangia, Amanpreet Singh, Julian Michael,
  Felix Hill, Omer Levy, and Samuel Bowman. 2019.
\newblock \href
  {https://proceedings.neurips.cc/paper/2019/file/4496bf24afe7fab6f046bf4923da8de6-Paper.pdf}
  {Superglue: A stickier benchmark for general-purpose language understanding
  systems}.
\newblock In \emph{Advances in Neural Information Processing Systems},
  volume~32. Curran Associates, Inc.

\bibitem[{Wang et~al.(2018)Wang, Singh, Michael, Hill, Levy, and
  Bowman}]{wang-etal-2018-glue}
Alex Wang, Amanpreet Singh, Julian Michael, Felix Hill, Omer Levy, and Samuel
  Bowman. 2018.
\newblock \href {https://doi.org/10.18653/v1/W18-5446} {{GLUE}: A multi-task
  benchmark and analysis platform for natural language understanding}.
\newblock In \emph{Proceedings of the 2018 {EMNLP} Workshop {B}lackbox{NLP}:
  Analyzing and Interpreting Neural Networks for {NLP}}, pages 353--355,
  Brussels, Belgium. Association for Computational Linguistics.

\bibitem[{Zhao et~al.(2020)Zhao, Lin, Mi, Jaggi, and Sch\"utze}]{mask}
Mengjie Zhao, Tao Lin, Fei Mi, Martin Jaggi, and Hinrich Sch\"utze. 2020.
\newblock \href
  {http://nbn-resolving.de/urn/resolver.pl?urn=nbn:de:bvb:19-epub-74038-6}
  {Masking as an efficient alternative to finetuning for pretrained language
  models}.
\newblock In \emph{The 2020 Conference on Empirical Methods in Natural Language
  Processing}.

\end{thebibliography}
\bibliographystyle{acl_natbib}




\end{document}